\documentclass[10pt,twocolumn,letterpaper]{article}

\usepackage{iccv}
\usepackage{times}
\usepackage{epsfig}
\usepackage{graphicx}
\usepackage{amsmath}
\usepackage{amssymb}

\usepackage{booktabs}
\usepackage{makecell}
\usepackage{multirow}
\usepackage{caption}
\usepackage{gensymb}
\usepackage[lined,boxed,commentsnumbered,ruled]{algorithm2e}
\usepackage{rotating}
\usepackage[caption=false]{subfig}

\usepackage[accsupp]{axessibility} 

\usepackage[colorlinks,breaklinks=true,bookmarks=false,linkcolor=blue]{hyperref}

\iccvfinalcopy 

\def\iccvPaperID{1982} 

\ificcvfinal\fi

\begin{document}

\title{PIT: Position-Invariant Transform for Cross-FoV Domain Adaptation}

\author{Qiqi Gu\textsuperscript{\rm 1}\footnotemark[1] \quad Qianyu Zhou\textsuperscript{\rm 1}\footnotemark[1] \quad
Minghao Xu\textsuperscript{\rm 1} \quad Zhengyang Feng\textsuperscript{\rm 1}  \\
Guangliang Cheng\textsuperscript{\rm 2 5 7} \quad Xuequan Lu\textsuperscript{\rm 3 $\dagger$} \quad Jianping Shi\textsuperscript{\rm 2} \quad Lizhuang Ma\textsuperscript{\rm 1 4 6 $\dagger$}\\
\textsuperscript{\rm 1}Shanghai Jiao Tong University, \textsuperscript{\rm 2}SenseTime Group Research\\
\textsuperscript{\rm 3}Deakin University,
\textsuperscript{\rm 4}East China Normal University, \textsuperscript{\rm 5}Shanghai AI Laboratory \\ \textsuperscript{\rm 6}MoE Key Lab of Artificial Intelligence, SJTU, 
\textsuperscript{\rm 7}Qing Yuan Research Institute, SJTU\\
{\tt\small \{miemie, zhouqianyu, xuminghao118, zyfeng97\}@sjtu.edu.cn}, \tt\small guangliangcheng2014@gmail.com \\
\tt\small xuequan.lu@deakin.edu.au,
\tt\small shijianping@sensetime.com,
\tt\small ma-lz@cs.sjtu.edu.cn
}

\maketitle
\ificcvfinal\thispagestyle{empty}\fi

\begin{abstract}
Cross-domain object detection and semantic segmentation have witnessed impressive progress recently. Existing approaches mainly consider the domain shift resulting  from external environments including the changes of background, illumination or weather, while distinct camera intrinsic parameters appear commonly in different domains and their influence for domain adaptation has been very rarely explored. 
In this paper, we observe that the Field of View (FoV) gap induces noticeable instance appearance differences between the source and target domains. 
We further discover that the FoV gap between two domains impairs domain adaptation performance under both the FoV-increasing (source FoV $\textless$ target FoV)
and FoV-decreasing cases. Motivated by the observations, we propose the \textbf{Position-Invariant Transform} (PIT) to better align images in different domains.  We also introduce a reverse PIT for mapping the transformed/aligned images back to the original image space, and design a loss re-weighting strategy to accelerate the training process. Our method can be easily plugged into existing cross-domain detection/segmentation frameworks, while bringing about negligible computational overhead. Extensive experiments demonstrate that our method can soundly boost the performance on both cross-domain object detection and segmentation for state-of-the-art techniques. 
Our code is available at \url{https://github.com/sheepooo/PIT-Position-Invariant-Transform}.
\end{abstract}
\section{Introduction}

\begin{figure}[t]
\centering
\includegraphics[scale=0.48]{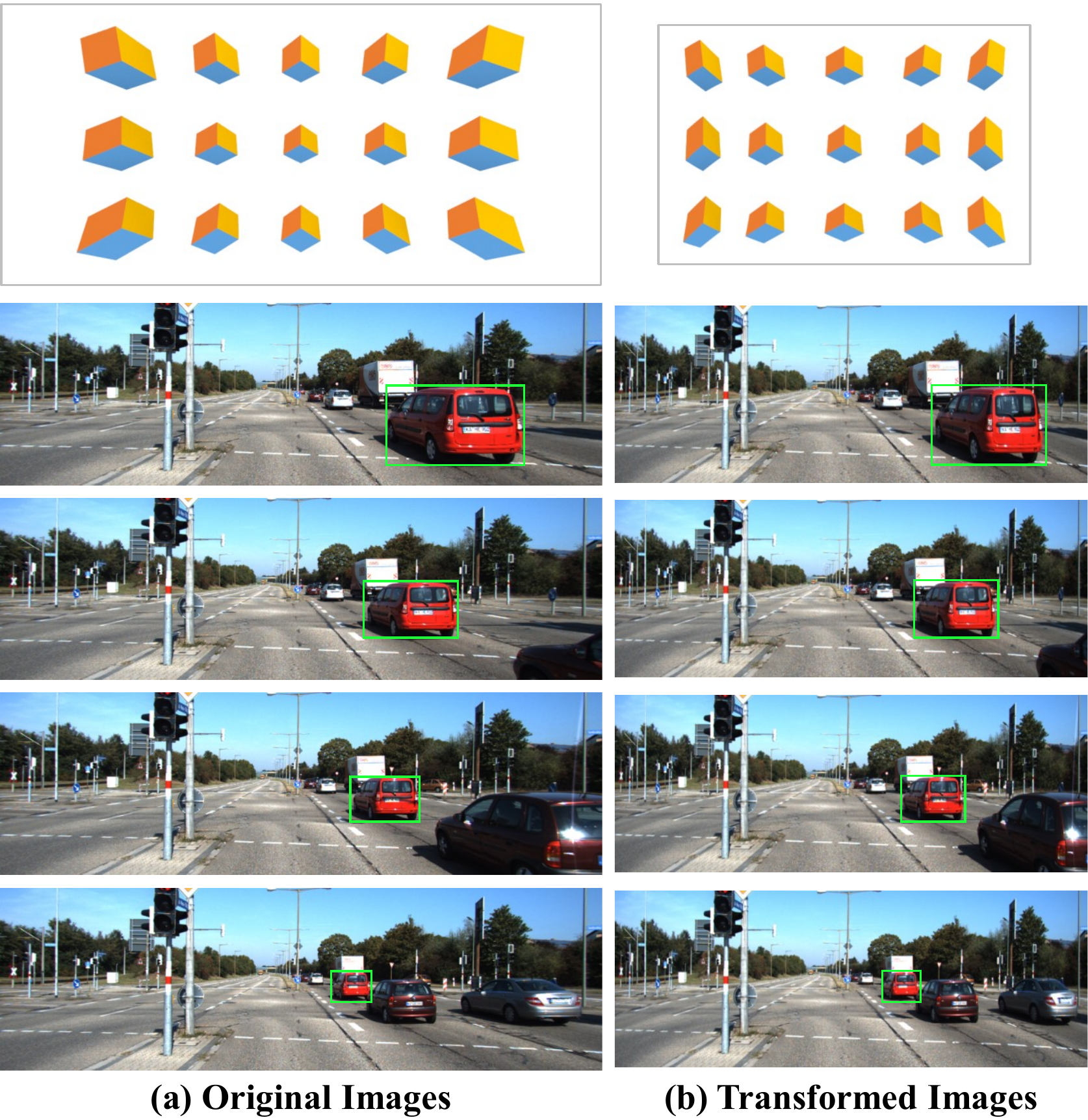}
\caption{Objects (cars) in different positions relative to the camera have different extent of deformation, which remarkably burdens the alignment of intra-class features. This can be effectively mitigated by our PIT. Top row: images of an object (in different positions) captured by a virtual camera.
Other rows: real photos from the KITTI dataset. 
}
\label{pit_compare}
\end{figure}

Object detection~\cite{girshick2015fast,FASTER-RCNN,yolo} and semantic segmentation~\cite{fcn,chen2017deeplab,chen2018encoder,feng2020semi} are two fundamental problems in computer vision. The former aims at precisely locating and identifying the objects in an image and the latter targets to classify the semantics of each pixel.\footnotetext{*Equal Contribution. $^\dagger$ Joint Corresponding author.} Training a generalized model with high performance for the two tasks calls for massive images with elaborate annotations, while it is laborious to prepare such well-annotated data. Meanwhile, due to the existence of domain shift~\cite{da_theory}, a model trained on a specific dataset often suffers from significant performance degradation when applied to another domain. A common solution is to transfer the knowledge acquired from a labeled source domain to an unlabeled target domain, which is known as Unsupervised Domain Adaptation (UDA)~\cite{da_survey}.

In general, two typical manners have been explored to adapt models from the source to the target domain. One is pixel-level alignment, target-like images are generated to provide implicit or explicit supervisory signals on target domain~\cite{gta,CyCADA,domain_mixup}. The other is feature-level alignment, the feature distributions of two domains are aligned through constraining domain discrepancy metrics~\cite{dan,deepcoral,wdan} or performing feature confusion~\cite{revgrad,adda,multi_adversarial}.

In the study of cross-domain detection/segmentation, previous works~\cite{DA-Faster-RCNN,SWDA,GPA,MTOR,choi2019self,ICR-CCR,CyCADA,AdaptSegNet,BDL,zhou2021context} mainly focus on narrowing the domain shift caused by external environments, \emph{e.g.} the change of background, illumination and weather, etc. However, very little attention has been paid to the camera's intrinsic parameters which often bring noticeable domain discrepancy due to the use of various cameras.

We observe that one main camera parameter, the Field of View (FoV)\textsuperscript{\rm 1}\footnotetext{\textsuperscript{\rm 1}Field of View  (FoV): in photography, the angle between two rays passing through the perspective center (rear nodal point) of a camera lens to the two opposite sides of the format. \label{fov_def}}, induces a distinct dimension of the domain gap. As a matter of fact, the FoV discrepancy frequently occurs among datasets or in real-world scenarios. 
For instance, in autonomous driving, cameras with different FoVs are often used together, because of the inevitable updating of cameras in the long period of data collection.
FoV difference derives the variety of instance structural appearances across the source and target domains, leading to the sample diversifying within a category. This obviously increases the burden of domain adaptation models, thus resulting in less desired performance. 

Motivated by the above observation, in this paper we attempt to alleviate the adverse impact of the diverse FoVs between domains, in order to boost the performance of cross-domain detection/segmentation.

We discuss the influence of the FoV gap in two general cases. (1) In FoV-increasing adaptation (the FoV of the target domain is larger than that of the source domain), the target domain instances with large incident angles cannot be well aligned to the source domain for the lack of similar-appearance counterparts. (2) In FoV-decreasing adaptation (target FoV smaller than source FoV), the sparsity of the source domain instances within a specific range of incident angle also hampers domain alignment. Existing UDA methods usually try to bridge the whole domain gap and optimize the model without specifically taking account of the FoV factor, thus preventing the model from fully learning  domain-invariant features. 

To address the above problem, we propose the \textbf{Position-Invariant Transform} (PIT) to straightforwardly narrow the FoV gap between the source and target domains (Fig. \ref{pit_compare}). Specifically, the pixels lying in the original imaging space are mapped to another two-dimensional space shaped as a spherical surface, such that the appearances of the instances in various positions are aligned to a great extent. Also, we introduce a reverse PIT for mapping the transformed images back to the original image space.  In addition, we design an efficient loss re-weighting strategy to speed up the training procedure. Our modules induce little computational overhead while boosting performance, and they can be easily served as plug-and-play modules to any existing cross-domain detection/segmentation frameworks.

Our contributions can be summarized as follows:
\begin{itemize}
    \item We statistically analyze the negative influence of FoV difference between the source and target domains on UDA models, in which both the increasing and decreasing of FoV between domains impair the domain alignment. 
    \item We propose the Position-Invariant Transform (PIT) to align instance structural appearances in different positions in each category, and reverse PIT to map the transformed images to the original image space. 
    We also introduce a loss re-weighting strategy to speed up the training procedure.
    \item The effectiveness of PIT is verified on both cross-domain detection and segmentation tasks. 
    Equipped with our modules, state-of-the-art UDA methods show soundly better performance than before. 
\end{itemize}

\section{Related Work}

\noindent\textbf{Unsupervised Domain Adaptation (UDA).} 
UDA aims to adapt the model trained on a labeled source domain to an unlabeled target domain by reducing the distribution gap between two domains. 
A group of recent approaches focused on minimizing the domain discrepancy~\cite{dan,deepcoral,wdan} metric (\emph{e.g.} Maximum Mean Discrepancy~\cite{2014Deep}), adversarial learning~\cite{revgrad,adda,multi_adversarial} or prototype-based alignment~\cite{semantic,minimax_entropy,GPA}. Despite the successes achieved in classification-based tasks~\cite{dan,revgrad,deepcoral,adda,gta,domain_mixup}, these methods work well on simple classification datasets (\emph{e.g.}  MNIST~\cite{lecun1998gradient} and
SVHN~\cite{netzer2011reading}), but can hardly be applied to more challenging tasks, \emph{e.g.} object detection and semantic segmentation.

\begin{figure*}[t]
\centering
\includegraphics[scale=0.7]{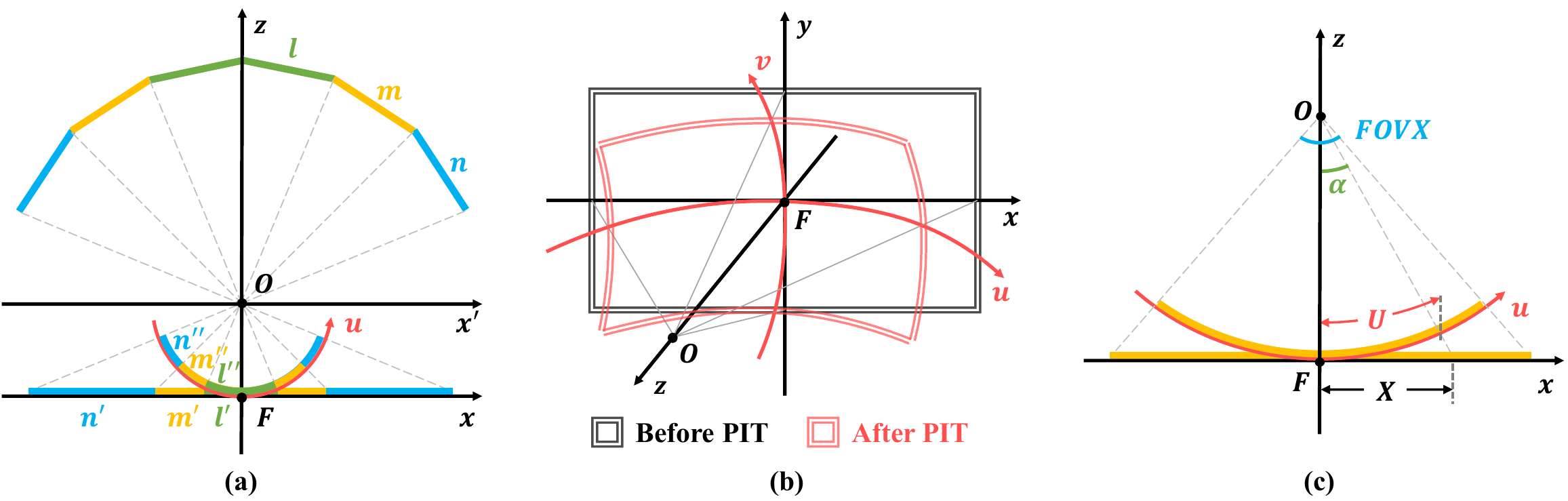}
\caption{\textbf{(a)} Illustration of the position-related deformation and the position-invariance of PIT. \textbf{(b)} The 3D spatial relationship of images before and after PIT. \textbf{(c)} The transformation between two coordinate systems. $O$: the optical center of a camera; $F$: the focal point; $x'Oy'$: the plane of lens ($y'$ axis is perpendicular to $x'Oz$); $xFy$: the imaging plane which is parallel to $x'Oy'$; $uFv$: the spherical surface to map the image, where the coordinate axes $u$ and $v$ are arcs. 
}
\label{fig:lens_imaging_2d}
\end{figure*}

\noindent\textbf{Domain Adaptive Detection/Segmentation.} Not until recently has the community paid attention to domain shift problem in object detection or semantic segmentation.
This line of research has been investigated by a large number of researchers, and great efforts have been made to explore a variety of algorithms and architectures to reduce the domain gap in pixel-level~\cite{CyCADA, DISE, BDL, LTIR,guo2021label}, feature-level~\cite{SIBAN, CBST,MTOR, DA-Faster-RCNN,zhou2021sad}, 
instance-level~\cite{DA-Faster-RCNN, ICR-CCR, HTCN} and output-level~\cite{AdaptSegNet,AdaptPatch,CLAN,APODA}, which have shown successes on both object detection~\cite{DA-Faster-RCNN, SCDA, MTOR,SWDA,HTCN,GPA,ICR-CCR} and semantic segmentation~\cite{CyCADA, AdaptSegNet, CLAN, BDL, SIM, AdaptPatch, ADVENT,zhou2021context}. The  current  mainstream  approaches of these two tasks include adversarial learning~\cite{SCDA,MAF,SWDA, AdaptSegNet,CLAN,ADVENT,AdaptPatch}, self-training~\cite{CBST, CRST,NL} and self-ensembling~\cite{MTOR,choi2019self,deng2020unbiased,zhou2020uncertainty,zhou2021context}. 
Despite the great progress, these works mainly focused on adapting different external environmental conditions, \emph{e.g.} background, illumination and weather. While the gap of camera intrinsic parameters between distinct domains has been ignored. 
In this work, we show the effectiveness of our method by easily integrating it into adversarial learning and self-ensembling on these two tasks.

\noindent\textbf{CNNs with Geometric Transformations. }
Researchers investigated CNNs with the abilities of geometric transformation or deformation gains over the past years. 
Spatial transformer networks \cite{STN} predicted the transformation parameters to reduce the influence of affine transformations. Active convolution \cite{ActiveConv} designed a transformable convolution kernel to get a more general shape of receptive field. Deformable convolution network \cite{DCN} further improved the former by predicting the receptive field location, and \cite{su2017learning} used spherical CNN to translate a planar CNN to process $360^\circ$ imagery directly in its equirectangular projection. Largely different from these methods which mainly focused on designing new network architectures, our method pays more attention to the attribution of the data itself (\emph{i.e.} position-related deformation caused by camera imaging) to enhance the feature alignment in UDA models.

\begin{figure*}[t]
\centering
\includegraphics[scale=0.45]{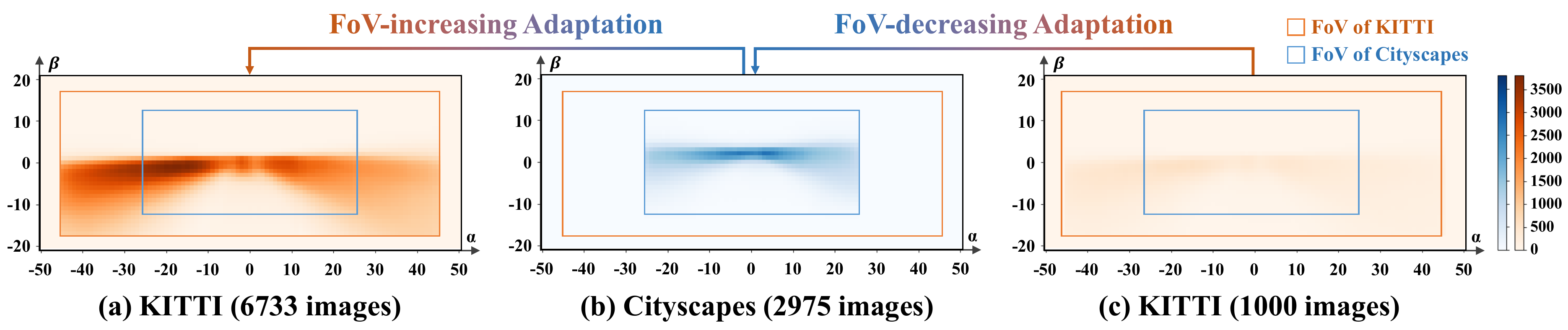}
\caption{Heat maps of foreground occurrences for each $(\alpha, \beta)$ in AP space. \textbf{(a)(c)} Statistics from the whole KITTI~\cite{kITTI} training set and a subset involving 1,000 images, which stand for large-scale and small-scale datasets in the real world, respectively. \textbf{(b)} Statistics from the Cityscapes training set. }
\label{fig:count}
\end{figure*}

\section{Method} 

In Unsupervised Domain Adaptation (UDA), a source domain $\mathcal{S} = \{(x^\mathcal{S}_i, y^\mathcal{S}_i)\}^{N_\mathcal{S}}_{i=1}$ with $N_\mathcal{S}$ labeled samples and a target domain $\mathcal{T} = \{x^\mathcal{T}_j\}^{N_\mathcal{T}}_{j=1}$ with $N_\mathcal{T}$ unlabeled samples are available, where $x^\mathcal{S}_i$ follows source distribution $\mathbb{P}_\mathcal{S}$, and $x^\mathcal{T}_j$ obeys target distribution $\mathbb{P}_\mathcal{T}$. The objective of UDA is to train a model generalizing well in the target domain, using the above data from both domains. 

\subsection{Motivation}
\label{sec:moti}

In the real world, images are often captured by cameras with distinct intrinsic parameters, which leads to the cross-camera domain gap. We observe that the structure of objects deform noticeably as their positions change, and the FoV parameter mainly impacts the deformation extent (Fig. \ref{pit_compare}). 

The FoV parameter restricts the angle of the area that can be observed by a camera, \emph{i.e.} the maximum incident angle of observable objects. Fig.~\ref{fig:lens_imaging_2d} (a) illustrates how the variance of the incident angle affects the structural appearance of an object. $l$, $m$, and $n$ are structure-alike objects which lie in different positions with the same distance to the optical center $O$. When projected onto the imaging plane, the length of their images $l'$, $m'$, and $n'$ are obviously different. Specifically, with the increase of an object's deviation from the center of a scene (\emph{i.e.} the expansion of the incident angle), its camera imaging becomes longer, which makes the object structure vary in different positions of an image.

Because of the restriction on the range of incident angles by FoV, the structural appearance of objects within the same category can be noticeably distinct between the source and target domains, as shown in Fig.~\ref{pit_compare} where different degrees of imaging deformation may occur in two domains. 
This kind of deformation is totally different from the lens distortion~\cite{distortion}. The latter is a deviation from rectilinear projection and can be fixed by camera calibration, and the calibrated image is the ideal projection on the imaging plane.

Considering the significance of learning structure-invariant feature representations in scene understanding~\cite{STN,DCN}, the structural difference between the objects from two domains can trap a UDA model into a dilemma in which that kind of difference cannot be handled well. To better elucidate the existence and underlying effect of the FoV gap between the two domains, we statistically analyze the incident angle distribution in various datasets. Specifically, we define $\alpha$ and $\beta$ ($\alpha$ is shown in Fig.~\ref{fig:lens_imaging_2d} (c), $\beta$ is the counterpart in $yFz$ plane) as a point's incident angles towards the optical center along the horizontal and vertical axis, respectively. Notice that the imaging deformation of an object is closely related to $\alpha$ and $\beta$, and the deformation extent continuously increases along these two angles' absolute values. Therefore, we span the $\alpha$ axis and $\beta$ axis to form a two-dimensional space, named as \textbf{Angular-Position space} (AP space), in which the absolute value of each point's coordinate measures the horizontal and vertical deformation extent of the object lying in the corresponding position. 
We then count the number of foreground occurrences for each $(\alpha, \beta)$ integer values on KITTI~\cite{kITTI} and Cityscapes~\cite{CITYSCAPES} datasets, and these statistics are displayed as heat maps in Fig.~\ref{fig:count}. 
It can be observed that the objects of KITTI dataset distribute in a wider range of incident angles than those of Cityscapes dataset, which leads to two opposite directions of cross-FoV adaptation (see below).

\noindent\textbf{FoV-increasing Adaptation.} In this case, the target domain possesses a wider FoV distribution than the source domain, \emph{e.g.} adapting from Cityscapes (Fig. \ref{fig:count}(b)) to KITTI (Fig. \ref{fig:count}(a)), which means that the objects in target domain own a greater range of deformation extents. Consequently, some target objects fall in the regions without source objects in the AP space, and they cannot be well aligned to the source domain for the lack of proper supervision from similar-appearance counterparts, which impairs the performance of UDA models. The proposed PIT module (Sec. \ref{sec:PIT}) effectively mitigates this defect via its position-invariance. 

\noindent\textbf{FoV-decreasing Adaptation.} In this case, the target domain has a narrower FoV distribution, \emph{e.g.} adapting from KITTI (Fig. \ref{fig:count}(a)(c)) to Cityscapes (Fig. \ref{fig:count}(b)), such that the distributional range of target objects are covered by that of source objects. It is true that when the source objects are dense enough everywhere (Fig. \ref{fig:count}(a)) in the AP space, domain alignment can be well performed by a UDA method. However, when the source domain has low data density (Fig. \ref{fig:count}(c)), a target object can hardly find its source counterparts with a similar structural appearance which it can align with; meanwhile the source samples are not fully utilized. Under this situation, the proposed PIT approach (Sec \ref{sec:PIT}) is able to gather source objects in the AP space and thus eases the alignment.
\label{section:FoV-decre}

\subsection{Position-Invariant Transform}
\label{sec:PIT}

The object deviating more from the principal axis of the lens would be stretched to a greater extent in the camera imaging process, which manifests the imaging deformation phenomenon in Fig.~\ref{pit_compare}.

In order to alleviate this kind of deformation, we propose the \textbf{Position-Invariant Transform} (PIT). Fig.~\ref{fig:lens_imaging_2d}(a) shows the principle of PIT. 
The location of a point's image is the intersection of its incident light passing through the optical center $O$ and the imaging surface, so the imaging of a scene would be altered by changing the imaging surface.
In this method, the incident light from an object passing through $O$ is received with a spherical surface instead of a plane, i.e. the $uFv$ surface with sphere center $O$ shown in Fig.~\ref{fig:lens_imaging_2d}(b). In such a spherical space, images can largely retain the relative size of original objects. For the same-size objects $l$, $m$ and $n$ in Fig.~\ref{fig:lens_imaging_2d}(a), they are mapped to $l''$, $m''$ and $n''$ with the same length on the $uFv$ surface. This example illustrates that the imaging on a spherical surface is invariant to the object's angular position, \emph{i.e.} satisfying \textit{position-invariance}.

\begin{figure*}[t]
\centering
\includegraphics[scale=0.43]{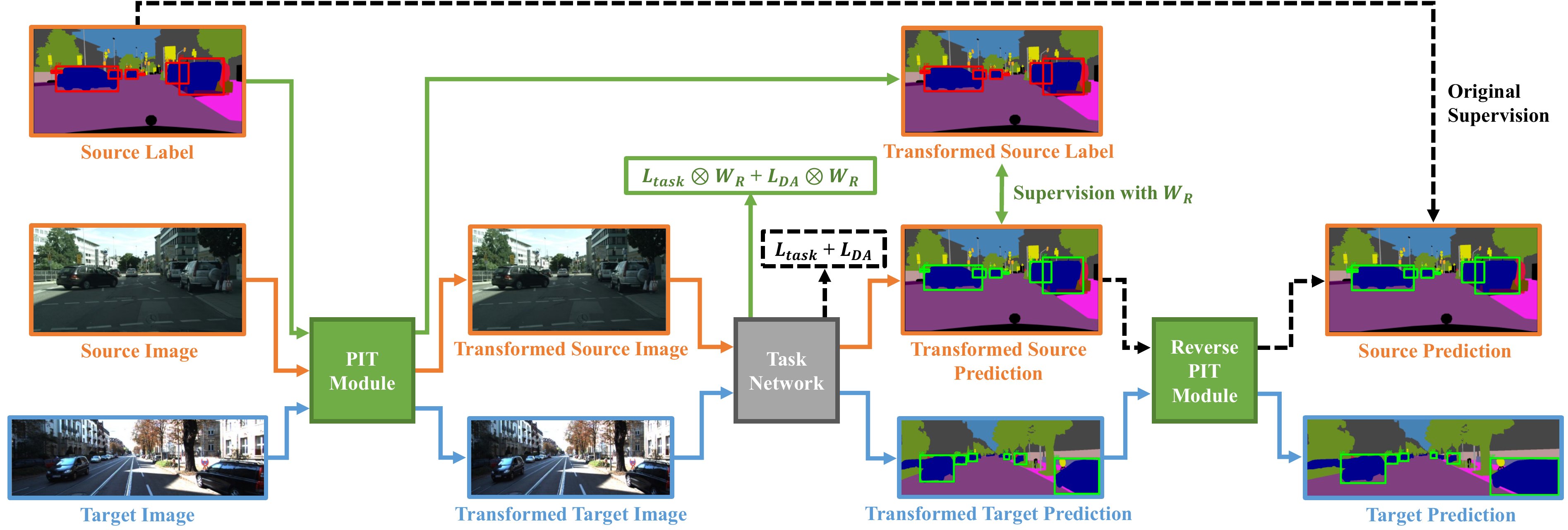}
\caption{ Overview of our method. 
}
\label{net}
\end{figure*}

After manifesting the benefits of a spherical surface over 2D imaging plane, 
a projection from spherical image back to a new plane image is needed to match the image with the input form of network.  
Thus, such projection should have two properties, which cannot be satisfied by the existing projection approaches (e.g. equirectangular, Mercator, etc.): (1) the image space after transformation should obey position-invariance, in order to align instances in the pixel level; (2) the horizontal (vertical) line should remain horizontal (vertical) after transformation, so as to ensure the validity of bounding box labels. Taking both properties into consideration, we formulate a new projection which is defined as (referring to Fig.~\ref{fig:lens_imaging_2d}(c) for intuitive notions): 

\begin{equation}
X(U) = f \times \tan(\frac{U}{f}),
\end{equation}
\begin{equation}
Y(V) = f \times \tan(\frac{V}{f}),
\end{equation}
\begin{equation}
M'[U][V] = M[X(U)][Y(V)],
\end{equation}
where $(X,Y)$ is the coordinate in the original image space (\emph{i.e.} the $xFy$ coordinate system with origin $F$), and $(U,V)$ is the coordinate in the image space after PIT (\emph{i.e.} the $uFv$ coordinate system with origin $F$). $M[X][Y]$ and $M'[U][V]$ denote the pixel values of the corresponding points before and after transformation. $f$ is the focal length which can be estimated using the FoV parameter or precisely calculated by camera calibration.


As  shown in Fig.~\ref{pit_compare}, the size of an image becomes smaller after PIT, and the regions further from the center of a scene are compressed with a higher ratio. Furthermore, the vertical/horizontal lines are preserved after PIT.

\subsection{Cross-FoV Domain Adaptation}
\label{sec:3.3}

\noindent\textbf{Integration.} The proposed PIT method can be utilized as a plug-and-play module to existing cross-domain detection and segmentation frameworks. As shown in Fig. \ref{net}, both the images from the source and target domains are first fed into the PIT module to be transformed into the position-invariant ones, which serve as the inputs to the task network. In the training phase, the labels from source domain are also transformed by PIT to provide supervision. As for inference, the prediction result of the task network is mapped back to the original image space by the reverse PIT module which outputs the final prediction.

\noindent\textbf{Reverse PIT and loss re-weighting strategy.} Since the evaluation is conducted with the un-transformed ground truth, it is plausible to provide supervision with the original labels, as shown by the black dash lines in Fig. \ref{net}. However, different from the PIT process which only needs to execute once for each input image in the datasets, the reverse PIT module would be employed in each iteration and cause extra computational cost. 
In order to accelerate the training, we design a pixel-wise loss re-weighting strategy to substitute the reverse PIT module during the training process. 
A pixel in the transformed image corresponds to a region in the original image, and each pixel in the original image weighs equally in evaluation. Therefore, a transformed pixel's weight should be the area of its mapping region, depending on the pixel's position. With this weight, the transformed supervision is equivalent to the reverse PIT in terms of loss computation. 
The weighting matrix is formulated as:
\begin{equation}
    w_{R}(U,V) = (X(|U| + 1) - X(|U|))\times(Y(|V|+1) - Y(|V|)),
\end{equation}
where $w_{R}$ is the weight assigned to pixel located in $(U,V)$ in the transformed image.

Using the weights derived above, we re-weigh the pixel-wise losses, including the task-specific loss $L_{task}$ (\emph{e.g.} the supervised loss $L_{sup}$ in \cite{choi2019self}) and the domain adaptation loss $L_{da}$ (\emph{e.g.} the consistency loss $L_{con}$ in \cite{choi2019self}):
\begin{equation}
    L = L_{task} \otimes W_{R} + \lambda L_{da} \otimes W_{R},
\end{equation}
where $\lambda$ is the weight to balance the two losses. 

With this loss re-weighting strategy, we can use the transformed labels to optimize the model, as shown by the green line in Fig.~\ref{net}, which speeds up the training procedure.

\section{Experiments} 
We conduct extensive experiments on object detection and semantic segmentation tasks. The results show that our approach can soundly boost the performance on cross-FoV adaptation by easily plugging it into any UDA frameworks.
\subsection{Experimental Setup}

\noindent\textbf{Datasets.} We utilize three public datasets provided with FoV parameters in our experiments: Cityscapes \cite{CITYSCAPES}, KITTI \cite{kITTI} and Virtual KITTI \cite{VKITTI}. 
In here, we add a two-dimensional array after the name of each dataset, to indicate the approximate horizontal and vertical FoV parameters $(FoVx, FoVy)$ of the camera for scene capturing.

\begin{itemize}

    \item \noindent\textbf{Cityscapes} \cite{CITYSCAPES} ($50\degree$, $26\degree$) is a dataset of street scenes in several cities. It owns 2,975 images for training and 500 for validation, and both of them have dense pixel-level labels. We get the bounding box labels for object detection task by calculating the tightest rectangles of instance annotations as \cite{DA-Faster-RCNN} did. 
    It uses 4 types of cameras with different FoVs ($49.5\degree < FoVx < 51.7\degree$, $25.5\degree < FoVy < 26.2\degree$), and we process each image with its own recorded FoV.

    \item \noindent\textbf{KITTI} \cite{kITTI} ($90\degree$, $34\degree$) is a real-world dataset containing 7,481 images with bounding boxes and another 200 images with pixel-level labels. In the detection task, we split the the training set and the validation set manually. 
    In the segmentation task, it is used as the target domain only due to the lack of pixel-level annotations. 

    \item \noindent\textbf{Virtual KITTI} \cite{VKITTI} ($80\degree$, $29\degree$) is a synthetic dataset which clones the scenes from the KITTI with 21,260 images. It provides pixel-level instance labels, and the bounding boxes are obtained as those in Cityscapes.

\end{itemize} 

\noindent\textbf{Baselines and Comparison Methods.} Following the experimental design in \cite{ICR-CCR}, we select SWDA~\cite{SWDA}, SCL~\cite{SCL}, GPA~\cite{GPA} as our baseline methods for cross-domain detection, and Self-Ensembling~\cite{choi2019self}, CowMix\cite{french2020milking}, CutMix\cite{french2019semi}, DACS~\cite{dacs} for cross-domain segmentation. 

We re-implement these methods for fair comparisons, and our re-implementations attain higher accuracies than the reported ones. When comparing with other state-of-the-art methods, we use the results from the original papers.

\noindent\textbf{Implementation Details.}
In object detection experiments, VGG16~\cite{vgg} model pre-trained on ImageNet \cite{imagenet} is used as the backbone of all the selected methods. 
The hyper-parameters are set according to the original papers. The average precision (AP) is used as evaluation metric.

In semantic segmentation experiments, the DeepLab-v2~\cite{chen2017deeplab} with ResNet101~\cite{he2016deep} pretrained on ImageNet~\cite{imagenet} and on MSCOCO~\cite{COCO} is used as our backbone. Hyper-parameters are set following \cite{AdaptSegNet,choi2019self}. 

\begin{table}[t]
  \centering
  \caption{Source-only detection results (car  AP, $\%$) traind on KITTI-50\degree  and tested on different degrees ($FoVx$) of cropped KITTI.}
    \begin{tabular}{c|cccc}
    \toprule
    FoVx & 50\degree   & 70\degree    & 80\degree   & 90\degree\\
    \midrule
    FR \cite{FASTER-RCNN}  & 87.49\% & 86.80\% & 86.31\% & 84.92\% \\
    FR + PIT & \textbf{87.81\%} &\textbf{ 87.37\% }& \textbf{86.88\%} & \textbf{86.43\%} \\

    \bottomrule
    \end{tabular}%
  \label{sourceonly}%
\end{table}%

\begin{figure}[t]
\centering
\includegraphics[scale=0.28]{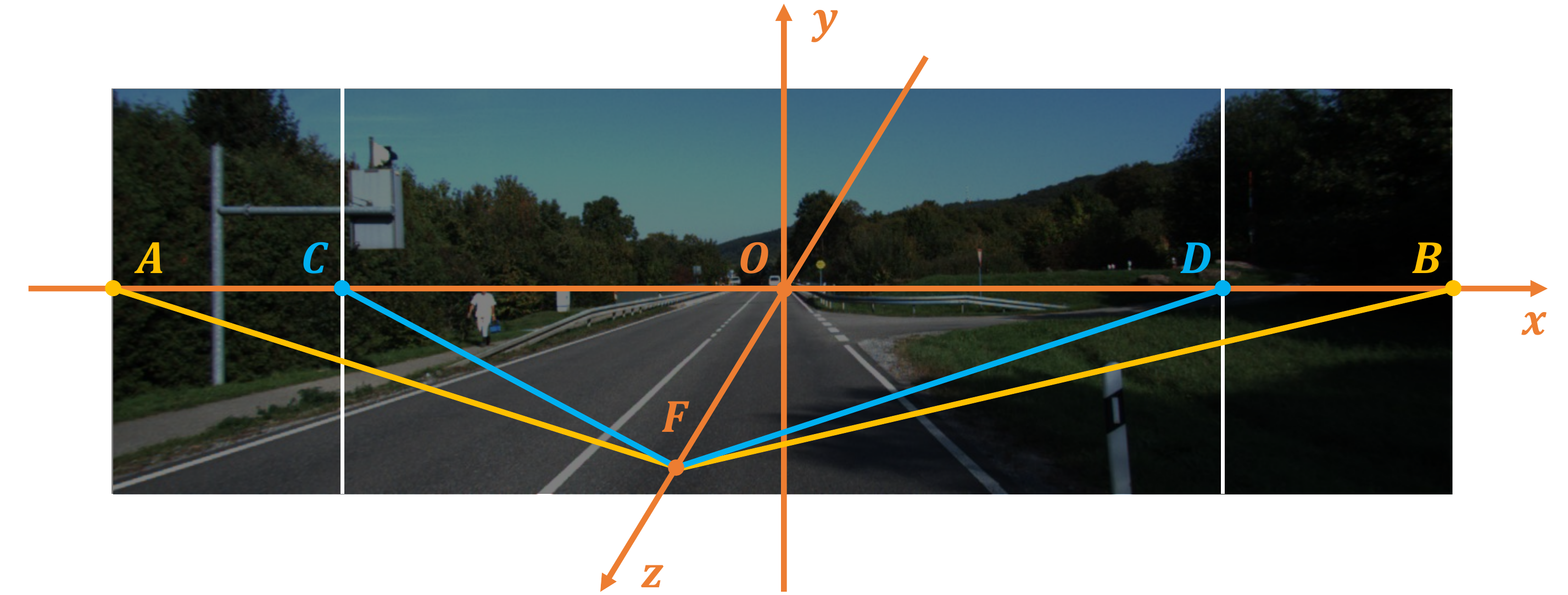}
\caption{Cropping image with certain $FoVx$. $FoVx$ was reduced from $\angle AFB$ to $\angle CFD$ after cropping.}
\label{fig:crop}
\end{figure}

\subsection{The Existence of FoV Gap}
In order to prove the existence of FoV gap, we crop the images (Fig. \ref{fig:crop}) to generate new datasets with certain $FoVx$. Then we train a Faster-RCNN \cite{FASTER-RCNN} model (but \textbf{NOT} a UDA method) on KITTI-50\degree, and test it directly on KITTI-70\degree/80\degree/90\degree to examine the compactness of features.

Tab. \ref{sourceonly} shows the detection results of these source-only experiments. Without PIT, the performance gets worse as the FoV gap gets bigger, while PIT effectively suppresses the performance drop. It demonstrates that the PIT module plays an important role in bridging the FoV gap.

\subsection{Domain Adaptation for Object Detection}

\begin{table*}[t]
  \centering
  \caption{Detection results of Virtual KITTI $\rightarrow$ KITTI (cropped with specified $FoVx$.)}
    \begin{tabular}{c|l|r|c|c|c|c|c}
    \toprule
    \multicolumn{1}{c|}{\multirow{2}[4]{*}{\shortstack{Source\\$FoVx$}}} & \multicolumn{1}{c|}{\multirow{2}[4]{*}{Method}} & \multicolumn{2}{c|}{Target $FoVx$: 50\degree} & \multicolumn{2}{c|}{Target $FoVx$: 70\degree} & \multicolumn{2}{c}{Target $FoVx$: 90\degree} \\
\cmidrule{3-8}          &       & car AP(\%) & Gain(\%)  & car AP(\%) & Gain(\%)  & car AP(\%) & Gain(\%) \\
    \midrule
    \multirow{2}[2]{*}{40\degree} & SCL (Arxiv'20)~\cite{SCL}   & \multicolumn{1}{c|}{69.81} & \multirow{2}[2]{*}{\textbf{0.75}} & 68.04 & \multirow{2}[2]{*}{\textbf{1.04}} & 65.71 & \multirow{2}[2]{*}{\textbf{3.33}} \\
          & SCL + PIT & \multicolumn{1}{c|}{\textbf{70.56}} &       & \textbf{69.08} &       & \textbf{69.04} &  \\
          
    \bottomrule
    \end{tabular}%
    
    \begin{tabular}{c|l|r|c|c|c|c|c}
    \toprule
    
        \multicolumn{1}{c|}{\multirow{2}[4]{*}{\shortstack{ Target\\$FoVx$}}} & \multicolumn{1}{c|}{\multirow{2}[4]{*}{Method}} & \multicolumn{2}{c|}{Source $FoVx$: 40\degree} & \multicolumn{2}{c|}{Source $FoVx$: 60\degree} & \multicolumn{2}{c}{Source $FoVx$: 80\degree} \\
\cmidrule{3-8}          &       & car AP(\%) & Gain(\%)  & car AP(\%) & Gain(\%)  & car AP(\%) & Gain(\%) \\
    \midrule
    \multirow{2}[2]{*}{90\degree} & SCL (Arxiv'20)~\cite{SCL}   & \multicolumn{1}{c|}{65.71} & \multirow{2}[2]{*}{\textbf{3.33}} & 67.74 & \multirow{2}[2]{*}{\textbf{2.46}} & 70.50 & \multirow{2}[2]{*}{\textbf{1.41}} \\
          & SCL + PIT & \multicolumn{1}{c|}{\textbf{69.04}} &       & \textbf{70.20} &       & \textbf{71.91} &  \\
  
    \bottomrule
    \end{tabular}%
  \label{tab:v2k_compare}%
\end{table*}%

\begin{table}[t]
  \centering
  \caption{ Detection results of Cityscapes $\rightarrow$ KITTI.}
    \begin{tabular}{l|c|c}
    \toprule
    Methods & car AP(\%) & Gain(\%) \\
    \midrule
    DAFRCN$^{\rm *}$(CVPR'18) \cite{DA-Faster-RCNN} & 64.10 & - \\
    SWDA$^{\rm **}$(CVPR'19) \cite{SWDA}  & 71.00  & - \\
    MAF$^{\rm *}$(ICCV'19) \cite{MAF}  & 72.10 & - \\
    SCL$^{\rm *}$(Arxiv'19) \cite{SCL}  & 72.70 & - \\
    ATF$^{\rm *}$(ECCV'20) \cite{ATF}  & 73.50 & - \\
    \midrule
    SWDA (CVPR'19) \cite{SWDA} & 72.42 & \multirow{2}[2]{*}{\textbf{3.35}} \\
    SWDA + PIT & \textbf{75.77} &  \\
    \midrule
    SCL (Arxiv'19) \cite{SCL}  & 75.28 & \multirow{2}[2]{*}{\textbf{1.84}} \\
    SCL + PIT  & \textbf{77.11} &  \\
    \midrule
    GPA (CVPR'20) \cite{GPA}  & 69.24     & \multirow{2}[2]{*}{\textbf{5.27}} \\
    GPA + PIT  & \textbf{74.51}     &  \\
    \bottomrule
    \end{tabular}%
    
    $^{\rm *}$ reported from its original paper, and $^{\rm **}$ from \cite{ATF}.

  \label{tab:c2k}%
\end{table}%
\begin{table}[htbp]
  \centering
  \caption{Detection results of Virtual KITTI $\rightarrow$ KITTI.}
    \begin{tabular}{l|c|c}
    \toprule
    Methods & car AP(\%) & Gain(\%) \\
    \midrule
    SWDA (CVPR'19)~\cite{SWDA}   & 69.74 & \multirow{2}[2]{*}{\textbf{2.12}} \\
    SWDA + PIT & \textbf{71.86} &  \\
    \midrule
    SCL (Arxiv'19)~\cite{SCL}   & 70.50 & \multirow{2}[2]{*}{\textbf{1.41}} \\
    SCL + PIT & \textbf{71.91} &  \\
    \midrule
    GPA (CVPR'20)~\cite{GPA}  & 65.36     & \multirow{2}[2]{*}{\textbf{5.35}} \\
    GPA + PIT & \textbf{70.71}     &  \\
    \bottomrule
    \end{tabular}%
    
  \label{tab:v2k}%
\end{table}%
\begin{table}[htbp]
  \centering
  \caption{Comparison with data augmentation.}
    \begin{tabular}{ccc|cc}
    \toprule
    SWDA \cite{SWDA}  & Aug   & PIT   & car AP(\%) & Gain(\%) \\
    \midrule
    \checkmark     &       &       & 72.42 & - \\
    \checkmark      & \checkmark      &       & 74.74 & 2.32 \\
    \checkmark      &       & \checkmark      & 75.77 & 3.35 \\
    \checkmark      & \checkmark      & \checkmark      & 76.93 & 4.51 \\
    \bottomrule
    \end{tabular}%
  \label{tab:data_aug}%
\end{table}%

\subsubsection{FoV-increasing Adaptation}

\noindent\textbf{Cityscapes ($50\degree$, $26\degree$) $\rightarrow$ KITTI $(90\degree, 34\degree)$.} It's a cross-camera adaptation, in which FoV gap is one of the main components of the domain gap. Table \ref{tab:c2k} shows the AP results of the car class. With our proposed PIT method, all the methods performed much better than their vanilla versions. The highest gain reaches $5.27\%$, which is a remarkable improvement in object detection. 

\noindent\textbf{Virtual KITTI $(80\degree, 29\degree)$ $\rightarrow$ KITTI $(90\degree, 34\degree)$.} It's a synthetic-to-real adaptation in which FoV gap is a minor factor of domain gap. The results are shown in Table \ref{tab:v2k}.

In order to look into the factors which influence the effectiveness of our method, we design controlled experiments. We crop the images (see Fig. \ref{fig:crop}) with certain $FoVx$ and use them as the source or target domain.

In Table \ref{tab:v2k_compare}, experiments in the upper part have the same source $FoVx$ and incremental target $FoVx$ (\emph{i.e.}, incremental FoV gap), and those in the bottom part of Table \ref{tab:v2k_compare} have a constant target $FoVx$ with different source $FoVx$. With the fixed $FoVx$ in one domain, the larger $FoVx$ gaps result in worse performance in the baseline, while our method gains higher improvement. These results verify that our proposed method can effectively narrow the specific FoV gap.

\subsubsection{FoV-decreasing Adaptation}
As analyzed in Section \ref{section:FoV-decre}, in this case, our method works with insufficient labeled data. So we reduce the size of the source dataset manually for the experiment setting, with no special treatment on the target domain. 

\noindent\textbf{KITTI $(90\degree$, $34\degree)$ $\rightarrow$ Cityscapes $(50\degree$, $26\degree)$}. We use 1,000 labeled images in KITTI dataset as source data. Table \ref{tab:k2c} shows the detection results on Cityscapes, and our method outperforms baselines by $1.48\% \sim 2.01\%$ on car AP.

\noindent\textbf{Virtual KITTI $(80\degree$, $29\degree)$ $\rightarrow$ Cityscapes $(50\degree$, $26\degree)$}. We use the "clone" subset (2126 images) of Virtual KITTI as source data. As shown in Table \ref{tab:v2c}, our method achieves increases when plugged in all the baseline networks.

\subsection{Domain Adaptation for Semantic Segmentation}

We conduct two  experiments : 1) Cityscapes $(50\degree$, $26\degree)$ $\rightarrow$ KITTI $(90\degree, 34\degree)$, 2) Virtual KITTI $(80\degree, 29\degree)$ $\rightarrow$ KITTI$(90\degree, 34\degree)$. mIoUs are reported for comparisons. The class-wise IoUs are reported in the supplementary material.

The results are shown in Table \ref{tab:ss_total}. Assembled in four state-of-the-art domain adaptative semantic segmentation methods, our method improves the mIoUs by $1.06\%$ to $1.77\%$  compared to the original methods, which again demonstrates the effectiveness of the proposed method. 

\subsection{Comparison with Data Augmentation}
Though served as a fixed part before and after the network, PIT is totally different from data augmentation. Data augmentation processes data with random parameters in several directions to diversify samples, while PIT aims at the opposite purpose. It calculates the optimal transform directly and reduces the variety of intra-class instances, which is beneficial for the feature alignment.

\begin{table}
\caption{Detection results of FoV-decreasing case.}
\centering
\subfloat[KITTI subset $\rightarrow$ Cityscapes.]
{
    \begin{tabular}{l|c|c}
    \toprule
    Methods & car AP(\%) & Gain(\%) \\
    \midrule
    SWDA (CVPR'19)~\cite{SWDA}  & 39.67 & \multirow{2}[2]{*}{\textbf{2.01}} \\
    SWDA + PIT & \textbf{41.68} &  \\
    \midrule
    SCL (Arxiv'20)~\cite{SCL}   & 38.64 & \multirow{2}[2]{*}{\textbf{1.61}} \\
    SCL + PIT & \textbf{40.25} &  \\
    \midrule
    GPA (CVPR'20)~\cite{GPA}   & 44.77     & \multirow{2}[2]{*}{\textbf{1.48}} \\
    GPA + PIT  & \textbf{46.25}     &  \\
    \bottomrule
    \end{tabular}%
     \label{tab:k2c}%
}

\subfloat[Virtual KITTI subset $\rightarrow$ Cityscapes.]
{
    \begin{tabular}{l|c|c}
    \toprule
    Methods & car AP(\%) & Gain(\%) \\
    \midrule
    SWDA (CVPR'19)~\cite{SWDA}  & 37.53 & \multirow{2}[2]{*}{\textbf{1.42}} \\
    SWDA + PIT & \textbf{38.95} &  \\
    \midrule
    SCL (Arxiv'19)~\cite{SCL}   & 37.22 & \multirow{2}[2]{*}{\textbf{1.49}} \\
    SCL + PIT& \textbf{38.71} &  \\
    \midrule
    GPA (CVPR'20)~\cite{GPA}   & 44.56     & \multirow{2}[2]{*}{\textbf{1.00}} \\
    GPA + PIT & \textbf{45.56}     &  \\
    \bottomrule
    \end{tabular}%
 \label{tab:v2c}%
}
 \label{tab:od_dec}%
\end{table}%

\begin{figure}[t]
\centering
\includegraphics[scale=0.33]{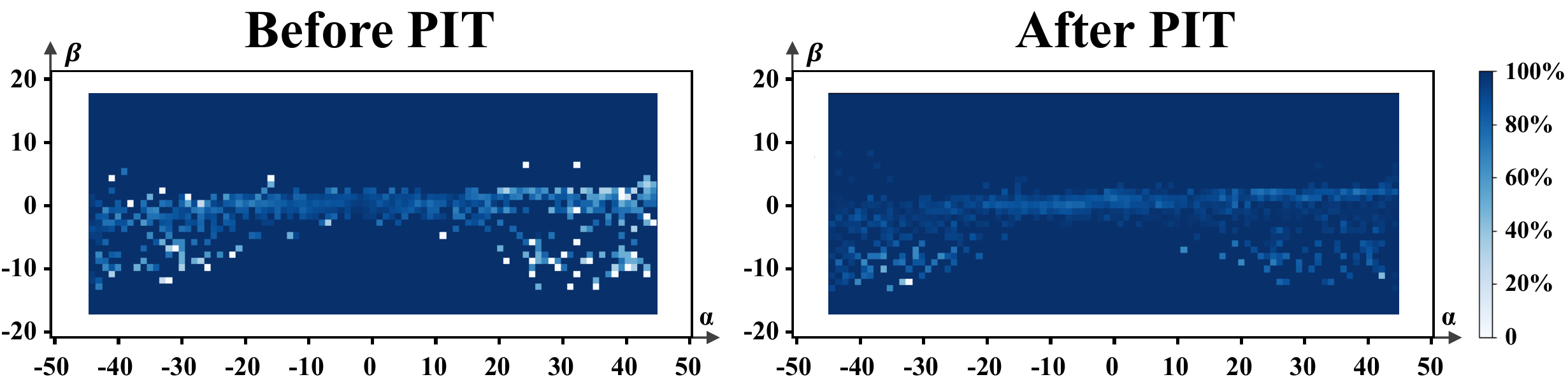}
\caption{The bin-wise performance in AP space.Based on detection task Cityscapes $\rightarrow$ KITTI, SWDA \cite{SWDA} backbone.}
\label{fig:bin-wise}
\end{figure}

\begin{table}
\caption{Segmentation results.}
\centering
\subfloat[Cityscapes $\rightarrow$ KITTI.]
{
    \begin{tabular}{l|c|c}
\toprule

Method                 &  mIoU
& Gain\\ 
\midrule

Self-Ensembling (ICCV'19) \cite{choi2019self}  &59.54 & \multirow{2}[2]{*}{\textbf{1.45}} \\
Self-Ensembling + PIT &\textbf{61.00} & \\
\midrule
CowMix (Arxiv'20) \cite{french2020milking}   & 59.15 & \multirow{2}[2]{*}{\textbf{1.22}} \\
CowMix + PIT & \textbf{60.37} \\
\midrule
CutMix (BMVC'20) \cite{french2019semi}   & 58.78 & \multirow{2}[2]{*}{\textbf{1.31}} \\
CutMix + PIT & \textbf{60.09} \\
\midrule
DACS (WACV'21) \cite{dacs}   & 59.19 & \multirow{2}[2]{*}{\textbf{1.63}} \\
DACS + PIT  & \textbf{60.82} \\

\bottomrule
\end{tabular}
}

\subfloat[Virtual KITTI $\rightarrow$ KITTI.]
{
    \begin{tabular}{l|c|c}
\toprule

Method                &  mIoU
& Gain\\ 
\midrule
GIO-Ada* (CVPR'19)~\cite{GIO_Ada}  	&53.5 & -\\
\midrule
Self-Ensembling (ICCV'19)~\cite{choi2019self}  	&55.45 & \multirow{2}[2]{*}{\textbf{1.77}} \\
Self-Ensembling + PIT 	&\textbf{57.22} \\
\midrule
CowMix (Arxiv'20)~\cite{french2020milking}   & 56.07 & \multirow{2}[2]{*}{\textbf{1.17}} \\
CowMix + PIT & \textbf{57.24} \\
\midrule
CutMix (BMVC'20)~\cite{french2019semi}   & 55.58 & \multirow{2}[2]{*}{\textbf{1.14}} \\
CutMix + PIT & \textbf{56.72} \\
\midrule
DACS (WACV'21)~\cite{dacs}   & 55.51 & \multirow{2}[2]{*}{\textbf{1.06}} \\
DACS + PIT & \textbf{56.57} \\

\bottomrule

\end{tabular}
}
\label{tab:ss_total}%
\end{table}%

We use the commonly-used data augmentation (random scale and random crop)~\cite{TrainInGermany} in experiments. Results in Tab. \ref{tab:data_aug} shows that PIT and data augmentation play different roles in the UDA task. Data augmentation aims at the linear transformation of objects (\emph{e.g.} different scale), while PIT reduces the instance diversity caused by non-linear deformations.

\subsection{Visualization}
In order to demonstrate the effectiveness of PIT over different incident angles, we reported the performance for different bins in the AP space (specified in Sec. \ref{sec:moti}). Using the center point to represent a predicted bounding box, we calculate the bin-wise accuracy and visualize in Fig. \ref{fig:bin-wise}. There are clear improvements in the peripheral regions where the objects have greater deformation, which verifies the effectiveness of the instance alignment through PIT. 
See more results in the supplementary material.

\section{Conclusion}
In this paper, we statistically analyzed the impact of FoV difference between domains, including both FoV-increasing and -decreasing cases.  Then we proposed a novel method (PIT) for cross-FoV detection/segmentation, which can be widely used in real-world applications due to the variety of cameras. 
Our method aligns the structural appearance of instances in the same category across domains. 
We also design a loss re-weighting strategy as a substitution of reverse PIT to speed up the training. 
As a plug-and-play approach, our method can be easily embedded into a wide range of existing networks. Experiments demonstrate that it boosts the performance in cross-domain detection and segmentation.
\section{Acknowledgement}
This work is supported by 
National Key Research and Development Program of China (No. 2019YFC1521104), 
Shanghai Municipal Science and Technology Major Project (No. 2021SHZDZX0102), 
Shanghai Science and Technology Commission (No. 21511101200),
Zhejiang Lab (No. 2020NB0AB01) 
and 
National Natural Science Foundation of China (No. 61972157).
The author Qianyu Zhou is supported by Wu Wenjun Honorary Doctoral Scholarship, AI Institute, Shanghai Jiao Tong University. 

\twocolumn[
\begin{@twocolumnfalse}
	\section*{\centering{PIT: Position-Invariant Transform for Cross-FoV Domain Adaptation
	\\Supplementary Material\\[90pt]}}
\end{@twocolumnfalse}
]

\section{Appendix.}

\subsection{Class-wise Detection Results}
In the paper, we follow the setting of DA-FRCN \cite{DA-Faster-RCNN}, SCL \cite{SCL} and GPA \cite{GPA} for fair comparison, which only have car results in the adaptation between Cityscapes $\rightarrow$ KITTI. In order to verify the generalization ability of our method, we conduct an experiment training with 4-classes label (4 overlapped classes in these two datasets). Results in Tab. \ref{tab:od_multi} shows that PIT can work on multi-class training.

\subsection{Class-wise Segmentation Results}
Tab. \ref{tab:ss_c2k} and Tab. \ref{tab:ss_v2k} shows the IoU of each class in semantic segmentation experiments. The results demonstrate that PIT module tends to improve the performance of large objects, for the reason that their area spans a larger FoV and thus lead to a greater extent of intra-instance deformation in the original images.

\subsection{Full-size FoV-decreasing Adaptation}
Sec. 3.2 analyzes the different situation (\emph{i.e.} whether the source image is sufficient) in FoV-decreasing case, and Sec. 4.3.2 gives the result of insufficient source images (a subset of source dataset). For reference, Tab. \ref{tab:od_fullsize} shows the result of sufficient source images (the fullsize source dataset). In FoV-decreasing case, the PIT module works better when there are not enough source samples.

\section{Computational Overhead}

Datasets can be transformed and saved before training, and it takes little time to transform an image. For example, it takes 0.27s to process an image in Cityscapes ($2048 \times 1024$ pixels), and 0.06s for one from KITTI ($1242 \times 375$ pixels) with a Tesla V100 GPU.

Table \ref{tab:time_ss} shows the time comparison of segmentation task Cityscapes$\rightarrow$KITTI with and without PIT. Due to the fact the Self-Ensembling~\cite{choi2019self} is the repredentative method of the consistency regulaization~\cite{french2019semi,french2020milking,dacs,zhou2020uncertainty,zhou2021context}, we use \cite{choi2019self} as our backbone framework. It needs little additional time to train with PIT and reverse PIT modules. Using our re-weighting strategy, training time for each iteration declines due to the smaller sizes of transformed images, and the performance remain similar (mIoU = $60.62\%$ for reverse PIT and $61.00\%$ for re-weighting). Adding the fixed time of PIT process, the average time rises little in few iterations, and even becomes less in a large number of iterations. In addition, the inference time per image in this task changes from $0.081$s to $0.096$s when adding our method, which only costs $10.9$s extra time for the validation of $748$ images.

\section{Qualitative results}

We visualize the qualitative results of task Cityscapes~\cite{CITYSCAPES} ($50\degree$, $25\degree$) $\rightarrow$ KITTI~\cite{kITTI} $(90\degree, 34\degree)$.

\begin{figure*}[t]
\centering
\includegraphics[scale=0.9]{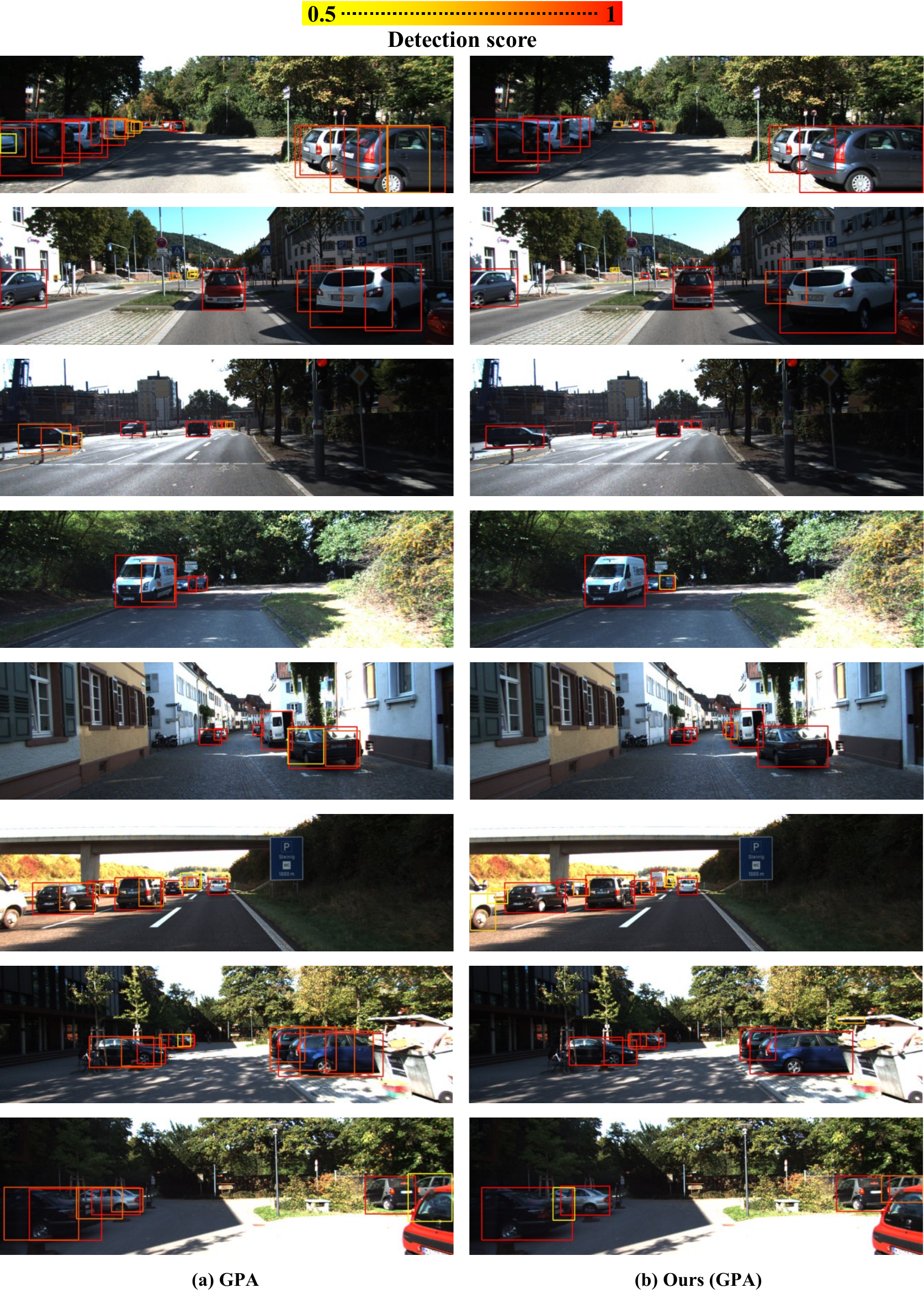}
\caption{Qualitative detection results of task Cityscapes $\rightarrow$ KITTI, all the predictions are car class.}
\label{fig:vis_od}
\end{figure*}

Fig.~\ref{fig:vis_od} shows the detection results using GPA~\cite{GPA} as the baseline. In results of the baseline (left column), the off-centered objects are likely to be recognized as several smaller objects or be detected partially due to their greater deformation extent. Our method (right column) solves these problem successfully by alleviating this kind of deformation, leading to clearer and more precise predicted bounding boxes.

\begin{figure*}[t]
\centering
\includegraphics[scale=1]{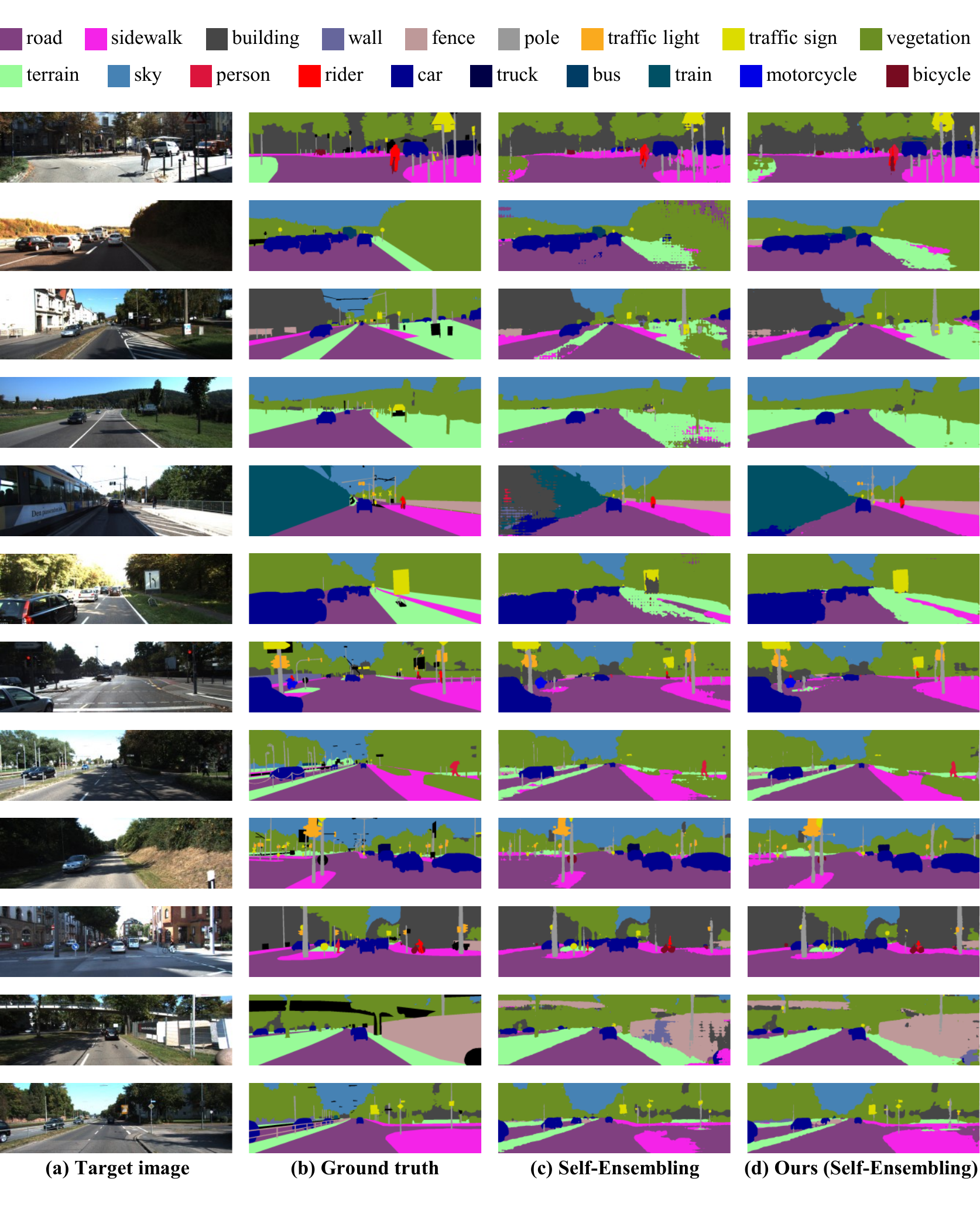}
\caption{Qualitative segmentation results of task Cityscapes $\rightarrow$ KITTI.}
\label{fig:vis_ss}
\end{figure*}

Using Self-Ensembling~\cite{choi2019self} as the baseline, we get the qualitative segmentation results in Fig.~\ref{fig:vis_ss}. Our method provides more accurate predictions, especially in the off-centered pixels.

\begin{table}[t]
  \centering
  \caption{Multi-class detection results (\%) of Cityscapes $\rightarrow$ KITTI.}
    \begin{tabular}{l|ccccc}
    \toprule
    Method & car   & person & rider & truck & mAP \\
    \midrule
    SWDA \cite{SWDA}  & 73.26 & 56.78 & 19.69 & 17.24 & 41.74 \\
    SWDA + PIT & \textbf{75.30} & \textbf{56.93} & \textbf{26.13} & \textbf{18.48} & \textbf{44.21} \\
    \bottomrule
    \end{tabular}%
  \label{tab:od_multi}%
\end{table}%

\begin{table}[t]
  \centering
  \caption{Detection results (carAP, $\%$) of KITTI $\rightarrow$ Cityscapes and Virtual KITTI $\rightarrow$ Cityscapes.}
    \begin{tabular}{l|cc}
    \toprule
    \multicolumn{1}{c|}{\multirow{2}[4]{*}{Method}} & \multicolumn{2}{c}{Datasets} \\
\cmidrule{2-3}          & K $\rightarrow$ C   & VK $\rightarrow$ C \\
    \midrule
    SWDA \cite{SWDA}  & 41.68 & 38.59 \\
    SWDA + PIT & \textbf{41.89} & \textbf{39.49} \\
    \bottomrule
    \end{tabular}%
  \label{tab:od_fullsize}%
\end{table}%

\begin{table*}[h]{}
\caption{Class-wise adaptive segmentation results (\%) of Cityscapes $\rightarrow$ KITTI.}
\label{table:c2k_seg}
\centering
\resizebox{\textwidth}{!}{%
\begin{tabular}{l|ccccccccccccccccccc|c}
\toprule
Method                 & \begin{turn}{90}road\end{turn} & \begin{turn}{90}sidewalk\end{turn} & \begin{turn}{90}building\end{turn} & \begin{turn}{90}wall\end{turn} & \begin{turn}{90}fence\end{turn} & \begin{turn}{90}pole\end{turn} & \begin{turn}{90}light\end{turn} & \begin{turn}{90}sign\end{turn} & \begin{turn}{90}vegetation\end{turn} & \begin{turn}{90}terrain\end{turn} & \begin{turn}{90}sky\end{turn} & \begin{turn}{90}person\end{turn} & \begin{turn}{90}rider\end{turn} & \begin{turn}{90}car\end{turn} & \begin{turn}{90}truck\end{turn} & \begin{turn}{90}bus\end{turn} & \begin{turn}{90}train\end{turn} & \begin{turn}{90}motocycle\end{turn} & \begin{turn}{90}bike\end{turn} & \begin{turn}{90}\textbf{mIoU$_{19}$}\end{turn}
\\ 
    \toprule
    Self-Ensembling~\cite{choi2019self}  & 85.36  & 48.30  & 80.50  & 37.98  & 39.95  & 45.64  & 58.95  & 53.08  & 87.86  & 52.70  & 93.40  & 58.09  & 47.01  & 87.21  & 52.20  & 68.08  & 37.41  & 49.24  & 48.28  & 59.54  \\
    Self-Ensembling + PIT & \textbf{88.86 } & \textbf{49.00 } & \textbf{81.18 } & \textbf{43.04 } & \textbf{40.90 } & 38.10  & 57.57  & \textbf{53.46 } & 87.43  & \textbf{57.15 } & \textbf{93.62 } & 52.93  & \textbf{47.75 } & \textbf{90.17 } & \textbf{60.52 } & \textbf{69.80 } & \textbf{65.04 } & 32.99  & \textbf{49.41 } & \textbf{61.00 } \\
    \midrule
    CowMix~\cite{french2020milking} & 85.26  & 49.16  & 80.64  & 38.63  & 41.93  & 43.16  & 60.47  & 56.97  & 86.81  & 47.65  & 93.08  & 58.41  & 42.23  & 87.15  & 49.16  & 66.58  & 40.66  & 46.80  & 49.09  & 59.15  \\
    CowMix + PIT & \textbf{90.44 } & 48.31  & \textbf{80.71 } & \textbf{41.46 } & 38.16  & 38.10  & 57.68  & 53.06  & \textbf{87.76 } & \textbf{60.81 } & \textbf{93.39 } & 50.27  & \textbf{50.87 } & \textbf{90.41 } & \textbf{58.05 } & \textbf{76.99 } & 36.95  & 42.48  & \textbf{51.13 } & \textbf{60.37 } \\
    \midrule
    CutMix~\cite{french2019semi} & 85.61  & 47.56  & 77.89  & 37.71  & 39.71  & 45.35  & 59.85  & 55.50  & 86.91  & 48.31  & 92.77  & 54.71  & 53.46  & 86.70  & 45.06  & 72.74  & 31.23  & 49.05  & 46.73  & 58.78  \\
    CutMix + PIT & \textbf{90.81 } & \textbf{48.40 } & \textbf{80.48 } & \textbf{48.99 } & 37.21  & 39.79  & 58.11  & 52.57  & \textbf{87.28 } & \textbf{61.70 } & 92.73  & 50.99  & 51.65  & \textbf{89.86 } & \textbf{52.67 } & 70.63  & \textbf{37.00 } & 42.56  & \textbf{48.20 } & \textbf{60.09 } \\
    \midrule
    DACS~\cite{dacs}  & 84.90  & 46.10  & 80.04  & 34.10  & 37.84  & 43.21  & 56.13  & 55.12  & 88.91  & 58.58  & 93.05  & 57.15  & 45.80  & 86.98  & 46.67  & 76.35  & 39.48  & 47.53  & 46.73  & 59.19  \\
    DACS + PIT & \textbf{90.30 } & \textbf{51.18 } & 78.63  & \textbf{41.22 } & \textbf{41.08 } & 41.35  & \textbf{59.29 } & \textbf{55.23 } & 86.75  & \textbf{59.15 } & 90.78  & 54.14  & \textbf{49.94 } & \textbf{88.63 } & \textbf{56.62 } & 66.15  & \textbf{56.67 } & 37.10  & \textbf{51.42 } & \textbf{60.82 } \\
    \bottomrule
\end{tabular}}
  \label{tab:ss_c2k}%
\end{table*}

\begin{table*}[h]{}
\caption{Class-wise adaptive segmentation results (\%) of Virtual KITTI $\rightarrow$ KITTI.}
\label{table:v2k_seg}
\centering
\begin{tabular}{l|cccccccccc|c}
\hline
Method                & \begin{turn}{90}road\end{turn} & \begin{turn}{90}building\end{turn} & \begin{turn}{90}pole\end{turn} & \begin{turn}{90}light\end{turn} & \begin{turn}{90}sign\end{turn} & \begin{turn}{90}vegetation\end{turn} & \begin{turn}{90}terrain\end{turn} & \begin{turn}{90}sky\end{turn}  & \begin{turn}{90}car\end{turn} & \begin{turn}{90}truck\end{turn} & \begin{turn}{90}\textbf{mIoU$_{10}$}\end{turn}
\\ 
\toprule
GIO-Ada*(CVPR'19)~\cite{GIO_Ada}  &81.4	&71.2	&11.3	&26.6	&23.6	&82.8	&56.5	&88.4	&80.1	&12.7	&53.5 \\
    \midrule
    Self-Ensembling~\cite{choi2019self} & 84.63  & 71.42  & 10.14  & 28.51  & 40.09  & 46.58  & 89.33  & 84.85  & 16.19  & 82.80  & 55.45  \\
    Self-Ensembling + PIT & \textbf{86.67 } & \textbf{72.83 } & 7.14  & \textbf{29.77 } & \textbf{40.70 } & \textbf{56.95 } & \textbf{90.13 } & \textbf{85.98 } & \textbf{16.95 } & \textbf{85.10 } & \textbf{57.22 } \\
    \midrule
    CowMix~\cite{french2020milking} & 83.89  & 68.96  & 12.58  & 30.30  & 39.02  & 50.77  & 89.05  & 84.02  & 18.00  & 84.14  & 56.07  \\
    CowMix + PIT & \textbf{87.16 } & 67.54  & 8.43  & \textbf{30.33 } & \textbf{42.53 } & \textbf{54.23 } & \textbf{91.10 } & \textbf{86.69 } & \textbf{20.85 } & 83.58  & \textbf{57.24 } \\
    \midrule
    CutMix~\cite{french2019semi} & 84.05  & 71.95  & 12.09  & 34.04  & 36.95  & 51.47  & 87.64  & 83.89  & 10.77  & 82.99  & 55.58  \\
    CutMix + PIT & \textbf{87.53 } & 66.08  & \textbf{12.54 } & 30.15  & \textbf{41.35 } & \textbf{55.61 } & \textbf{90.83 } & \textbf{86.56 } & \textbf{12.76 } & \textbf{83.79 } & \textbf{56.72 } \\
    \midrule
    DACS~\cite{dacs}  & 87.35  & 66.81  & 10.49  & 30.24  & 41.94  & 54.92  & 90.97  & 86.63  & 16.81  & 83.69  & 56.98  \\
    DACS + PIT & 85.82  & \textbf{72.21}  & 5.08  & 26.64  & \textbf{42.07 } & \textbf{52.87 } & 90.43  & \textbf{86.07 } & \textbf{20.50 } & \textbf{84.03 } & \textbf{56.57 } \\
    \bottomrule
\end{tabular}
\raggedright
$^{\rm *}$ the reported performance from its original paper.
  \label{tab:ss_v2k}%
\end{table*}

\begin{table*}[!h]
  \centering
  \caption{Time comparison of segmentation task Cityscapes $\rightarrow$ KITTI on a Tesla V100 GPU. RPIT refers to directly using reversed PIT in loss calculation, while re-weighting means using our proposed re-weighting strategy. }
    \begin{tabular}{c|l|cccc}
    \toprule
    Iteration & Method & $T_{PIT}$(s) & $T_{train}$ (s) & $T_{total}$(s) & $T_{average}$(s) \\
    \midrule
    \multirow{3}[2]{*}{10k} & Self-Ensembling~\cite{choi2019self} & 0     & 9,454.8  & 9,454.8  & 0.95  \\
          & Self-Ensembling + PIT (RPIT) & 1,207.2 & 9,849.1  & 11,056.3  & 1.11  \\
          & Self-Ensembling + PIT (re-weighting) & 1,207.2 & 9,150.3  & 10,357.5 & 1.04  \\
    \midrule
    \multirow{3}[2]{*}{100k} & Self-Ensembling~\cite{choi2019self} & 0     & 94,548.0  & 94,548.0  & 0.95  \\
          & Self-Ensembling + PIT (RPIT) & 1,207.2 & 98,491.0  & 99,698.2  & 1.00  \\
          & Self-Ensembling + PIT (re-weighting) & 1,207.2 & 91,503.0  & 92,710.2 & 0.93  \\
    \bottomrule
    \end{tabular}%
  \label{tab:time_ss}%
\end{table*}%

 \quad  

 \quad  
 
 \quad  
 
 \quad  
 
 \quad  
 
 \quad  
 
 \quad  

 \quad  
 
 \quad  
 
 \quad  
 
 \quad  
 
 \quad  
 
 \quad  
 
 \quad  
 
 \quad  
 
 \quad  
 
 \quad  
 
 \quad

{\small
\bibliographystyle{ieee_fullname}
\bibliography{egbib}
}

\end{document}